\documentclass[letterpaper]{article} 
\usepackage{aaai2026}  
\usepackage{times}  
\usepackage{helvet}  
\usepackage{courier}  
\usepackage[hyphens]{url}  
\usepackage{graphicx} 
\usepackage{natbib}  
\usepackage{caption} 
\usepackage{algorithm}

\usepackage{newfloat}
\usepackage{listings}
\DeclareCaptionStyle{ruled}{labelfont=normalfont,labelsep=colon,strut=off} 
\floatstyle{ruled}
\newfloat{listing}{tb}{lst}{}
\floatname{listing}{Listing}
\usepackage{booktabs}
\usepackage{algpseudocode}
\usepackage{amsmath}
\usepackage{amssymb}
\usepackage{subcaption}
\usepackage{multirow}
\usepackage{graphicx}
\usepackage{pifont}
\usepackage{listings} 
\usepackage{colortbl}
\usepackage{cleveref}
\usepackage{enumitem} 
\usepackage{amsthm}
\usepackage[most]{tcolorbox}
\usepackage{lipsum}
\usepackage{xurl}
\title{WaterMod: Modular Token-Rank Partitioning for Probability-Balanced LLM Watermarking}

\author{
    Shinwoo Park\textsuperscript{\rm 1},
    Hyejin Park\textsuperscript{\rm 2},
    Hyeseon Ahn\textsuperscript{\rm 1},
    Yo-Sub Han\textsuperscript{\rm 1}\thanks{Corresponding author.}
}
\affiliations{
    \textsuperscript{\rm 1}Yonsei University, Seoul, Republic of Korea\\
    \textsuperscript{\rm 2}Rensselaer Polytechnic Institute, Troy, NY, USA\\
    \{pshkhh, hsan, emmous\}@yonsei.ac.kr, parkh12@rpi.edu
}

\begin{document}

\maketitle

\newcommand{\eg}{{\it e.g.}}
\newcommand{\ie}{{\it i.e.}}

\begin{abstract}

Large language models now draft news, legal analyses, and software code 
with human-level fluency. 
At the same time, regulations such as the EU AI Act mandate that each synthetic passage 
carry an imperceptible, machine-verifiable mark for provenance.
Conventional logit-based watermarks satisfy this requirement by selecting a 
pseudorandom green vocabulary at every decoding step and boosting its logits, 
yet the random split can exclude the highest-probability token and thus erode fluency. 
WaterMod mitigates this limitation through a probability-aware modular rule. 
The vocabulary is first sorted in descending model probability; 
the resulting ranks are then partitioned by the residue $\text{rank}\bmod k$, 
which distributes adjacent—and therefore semantically similar—tokens across different classes.
A fixed bias of small magnitude is applied to one selected class. 
In the zero-bit setting~($k=2$), an entropy-adaptive gate selects either the even or the odd parity 
as the green list. 
Because the top two ranks fall into different parities, this choice embeds a detectable signal 
while guaranteeing that at least one high-probability token remains available for sampling.
In the multi-bit regime~($k>2$), 
the current payload digit $d$ selects the color class whose ranks satisfy 
$\text{rank} \bmod k = d$.
Biasing the logits of that class embeds exactly 
one base-$k$ digit—equivalently $\log_{2}k$ bits—per decoding step, 
thereby enabling fine-grained provenance tracing.
The same modular arithmetic 
therefore supports both binary attribution and 
rich payloads. 
Experimental results demonstrate that WaterMod consistently attains strong watermark detection 
performance while maintaining generation quality in both zero-bit and multi-bit settings. 
This robustness holds across a range of tasks, 
including natural language generation, mathematical reasoning, and code synthesis.
Our code and data are available at \url{https://github.com/Shinwoo-Park/WaterMod}.

\end{abstract}

\section{Introduction}
\label{sec:introduction}

Large language models~(LLMs) now draft news copy~\citep{goyal2022news}, 
answer legal queries~\citep{hu-etal-2025-fine}, 
and refactor production code~\citep{cordeiro2024empirical} with near‑human fluency~\citep{achiam2023gpt}. 
The same realism, however, obscures provenance and amplifies downstream risks—misinformation~\citep{pan2023risk}, 
plagiarism~\citep{lee2023language}, 
and data‑poisoned training corpora~\citep{sander2024watermarking,sander2025detecting}. 
Regulators have begun to respond.  
The European Union’s Artificial Intelligence Act~(EU AI Act) requires that outputs generated by 
general-purpose AI models be clearly identified as such, 
with disclosure obligations expected to take effect by 2026. 
Watermarking is among the recommended mechanisms for satisfying this requirement, 
aligning with principles already established for the 
disclosure of deepfake content~\citep{wilmerhale2024aiact}.
Regulatory guidance emphasizes that the disclosure mark should remain resilient to 
common post-processing operations and be verifiable through algorithmic means~\citep{euaiact2024}. 
Industry stakeholders are increasingly converging on watermarking as a practical solution.
OpenAI publicly acknowledges an internal text‑watermark detector 
under evaluation for ChatGPT~\citep{openai2024}, 
while Google DeepMind has released SynthID-Text~\citep{dathathri2024scalable}, 
a watermarking algorithm for LLM-generated text. 
These efforts reflect a growing consensus that imperceptible identifiers, 
such as watermarks, 
offer the most practical pathway toward 
regulatory compliance~\citep{golowich2024edit,hu2024inevitable,giboulot2024watermax,li2024segmenting,pang2024nofreelunch,zhou2024bileve,fu2024watermarking,panaitescu2025can}.

The dominant research thread biases sampling toward a pseudorandom green list of tokens 
while a detector counts 
their statistical over-representation.  
\citet{kirchenbauer2023watermark} randomly partition the vocabulary into a green list and a red list,
and force the decoder to prefer tokens from the green list at each decoding step.
While conceptually simple, such random partitioning frequently assigns contextually appropriate tokens 
to the forbidden set~(\ie, the red list), thereby reducing lexical diversity and harming fluency.
\citet{chen-etal-2024-watme} mitigate semantic degradation by introducing lexical-redundancy clusters, 
which help ensure that at least one suitable synonym remains in the green list. 
They achieve this by clustering synonyms using WordNet~\citep{fellbaum1998wordnet} 
look-ups or LLM prompting. 
However, the method relies on external synonym resources and prompt engineering, 
which introduces issues such as limited dictionary coverage, polysemy-related errors, 
and prompt sensitivity, ultimately hindering consistency across domains.
A parallel line of research~\citep{guo2024context} 
applies locality-sensitive hashing~(LSH) over token embeddings 
to induce semantically coherent partitions. 
While this approach enables clustering based on semantic similarity, 
it remains susceptible to collision errors and instability arising from hyperplane sensitivity, 
often leading to brittle behavior and semantic drift.
Moreover, most existing approaches 
rely on zero-bit watermarking, which merely indicates that a text has been generated 
by an AI model, without embedding any richer information to support provenance tracing. 
This limitation becomes critical in high-stakes applications—such as tracking leaked fine-tuning data or 
identifying the specific model instance responsible for generating disinformation—where regulators and 
service providers require more expressive payloads to ensure traceability.

We propose WaterMod, \textbf{Wa}termarking via \textbf{T}ok\textbf{e}n‑\textbf{r}ank \textbf{Mod}ular 
Arithmetic, 
a probability‑balanced watermarking framework that replaces 
heuristic green/red vocabularies with a 
modular partitioning of the vocabulary.
\begin{itemize}
    \item \textbf{Probability-ranked palette}: At each decoding step, the vocabulary is sorted in descending order 
    based on the conditional probabilities assigned by the model.
    Since tokens with contiguous ranks are deemed contextually similar and interchangeable by the model, 
    rank-based partitioning naturally preserves high-probability candidates.
    \item \textbf{Modular coloring}: Token ranks are partitioned 
    by \(\text{rank} \bmod k\): \(k = 2\) yields a zero-bit split, 
    while \(k > 2\) supports multi-bit embedding.
\end{itemize}
WaterMod builds entirely on the probability scores the model 
already produces, so it needs no synonym dictionaries, 
hashing tricks, or prompt engineering. 
By partitioning the ranked vocabulary into residue classes, 
WaterMod deliberately distributes near-synonyms across distinct color groups, 
ensuring that each decoding step retains a high-probability token within the designated class.
The result is a fluent yet verifiable watermark that can be 
flexibly scaled—from a binary attribution tag to a full multi-bit provenance 
string—positioning the method for future transparency and disclosure mandates.
Our experimental results show that WaterMod reliably embeds 
detectable watermarks without compromising content quality 
across a range of tasks—including natural language generation, 
mathematical reasoning, 
and code generation—in both zero-bit and multi-bit watermarking settings.

\section{Preliminaries}
\label{sec:preliminary}

\subsection{Text Generation in LLMs}

Let $\mathcal{V}$ be the vocabulary ($|\mathcal{V}|=V$) and
$x_{<t}=x_{1:t-1}$ the prefix available at time step~$t$.
An LLM with parameters~$\theta$ produces a
\emph{logit vector}
\[
  \boldsymbol{\ell}_{t}=f_\theta(x_{<t})\in\mathbb{R}^{V},
\]
where each entry $\ell_{t,i}$ denotes the unnormalized compatibility score 
for token $v_i\!\in\!\mathcal{V}$.
Sampling probabilities are defined by the softmax function:
\begin{equation}
  p_\theta\!\bigl(x_t=v_i \mid x_{<t}\bigr)
  \;=\;
  \frac{\exp(\ell_{t,i})}{\sum_{j=1}^{V}\exp(\ell_{t,j})}.
  \label{eq:softmax}
\end{equation}
The decoder draws $x_t$ from equation~\ref{eq:softmax},
appends it to the context, and repeats until an end‑of‑sequence token is produced.

\subsection{Logit-based Text Watermarking}

\paragraph{Embedding.}
Logit‑based watermarking perturbs $\boldsymbol{\ell}_{t}$ before sampling.
At each step a \emph{green list} $\mathcal{G}_t\subset\mathcal{V}$
and \emph{red list} $\mathcal{R}_t=\mathcal{V}\setminus\mathcal{G}_t$
are determined~(typically via a hash seeded by the previous token).
Given a bias magnitude $\delta>0$, the encoder raises the logits of green tokens:
\[
  \tilde{\ell}_{t,i}
  =
  \begin{cases}
    \ell_{t,i}+\delta, & v_i\in\mathcal{G}_t,\\
    \ell_{t,i},        & v_i\in\mathcal{R}_t.
  \end{cases}
\]
Sampling from $p_\theta\!\bigl(x_t=v_i \mid x_{<t}\bigr)$ 
is thus biased toward $\mathcal{G}_t$; 
a position is considered watermarked if the generated token at that step falls within $\mathcal{G}_t$.

\paragraph{Detection.}
Let a generated sequence have length $T$ and let
$G=\#\{\text{green tokens}\}$.
Under the null hypothesis $\mathcal{H}_{0}$~(no watermark, i.e.\ $\delta=0$),
each position is green with a fixed probability
\[
  \varepsilon \;=\; \frac{\lvert\mathcal{G}_t\rvert}{V}
  \quad (\text{assumed identical for all }t),
\]
so the total green count $G$ follows a binomial distribution: $G \sim \operatorname{Binom}(T,\varepsilon)$.
The detector evaluates the $z$‑score
\begin{equation}
  z \;=\;
  \frac{G - T\varepsilon}{\sqrt{T\varepsilon(1-\varepsilon)}},
  \label{eq:zscore}
\end{equation}
which converges to the standard normal distribution under $\mathcal{H}_{0}$ and
shifts to larger values when $\delta>0$~(\ie, when a watermark is embedded).
A one‑sided hypothesis test is performed: if $z$ exceeds a pre‑specified
threshold $\tau$~(chosen to attain the desired false‑positive rate),
$\mathcal{H}_{0}$ is rejected and the sequence is declared watermarked.

\subsection{Modular Arithmetic}

For an integer modulus $k\!\ge\!2$, the set of integers decomposes into
residue classes
\[
  [r]_k
  \;=\;
  \bigl\{\,n\in\mathbb{Z}\mid n\equiv r \pmod{k}\bigr\},
  \quad
  r\in\{0,\dots,k-1\}.
\]
A key advantage of using modular arithmetic lies in its collision-free and deterministic partitioning property. 
Each token rank~(logit-sorted rank) is deterministically mapped to a unique residue class, 
yielding nearly uniform class sizes that differ by at most one. 
This partitioning scheme requires no external resources such as lexicons or hash functions, 
making it efficient to implement.

\section{Methods}
\label{sec:methods}

At each decoding step, tokens are first sorted in descending order of model probability.  
In the \emph{zero-bit} setting, 
the vocabulary is partitioned 
into even- and odd-ranked tokens,  
which are alternately assigned to the green and red groups.  
When the distribution is sharp—\ie, 
most probability mass lies on the top one or two tokens—the entropy gate assigns a low probability 
to the odd-rank choice, 
so the even-ranked group is more likely to become green.
As the distribution flattens and entropy rises, 
that probability increases, making the odd-ranked group increasingly likely to be selected instead. 
This dynamic assignment ensures that at least 
one high-probability token remains in the green set,  
preserving fluency while enabling watermark insertion.  
Next, a small bias is added to the logits of green tokens 
to subtly guide sampling.  
During detection, the presence of a watermark is inferred by testing 
whether green tokens are statistically overrepresented.

In the \emph{multi-bit} setting, 
the parity rule is extended to 
$\text{rank}\bmod k$, which partitions the probability-sorted vocabulary
into $k$ color classes.  
A pseudorandom function~(PRF) permutes the payload digits, and at each
decoding step the hash of the previous token selects 
a target position $p$.  
The current base-$k$ digit $m=\mathbf{m}[p]$ then determines the class
with indices satisfying $\text{rank}\bmod k=m$, and only the logits of
those tokens receive the bias~$\delta$.  
Each generated token therefore carries one base-$k$ digit, 
\ie\ $\log_2 k$ payload bits in expectation.  
At detection time, majority voting over the observed color counts
recovers every digit; the repeated observations act as a natural form
of error correction and maintain robustness even for short passages.

\subsection{Zero-bit Watermarking}
\label{sec:zero_bit}

\paragraph{Probability-sorted parity partition.}
Algorithm~\ref{alg:zerobit_embedding} describes the watermark 
embedding procedure of WaterMod in zero-bit case. 
Given logits \(\boldsymbol{\ell}_{t}\) the permutation
\(\pi=\operatorname*{argsort}(\boldsymbol{\ell}_{t};\downarrow)\)
orders the vocabulary by the model probability.
Mapping \(r\mapsto r\bmod2\) creates disjoint even and odd classes.
Adjacent ranks, which the model views as interchangeable, are
distributed across the two classes.


\begin{algorithm}[hbt!]
\caption{\textsc{Zero-bit}\,: Embedding at step $t$}
\label{alg:zerobit_embedding}
\small
\textbf{Input:} logits $\boldsymbol{\ell}_{t}\!\in\!\mathbb{R}^{V}$, previous token $x_{t-1}$, secret key $K$, entropy scaling factor $H_{\text{scale}}$, bias $\delta$ \\
\textbf{Output:} next token $\hat{x}_{t}$
\begin{algorithmic}[1]
\State $p_i\!\gets\!\operatorname{softmax}(\boldsymbol{\ell}_{t})_i$
\State $H_t \gets -\displaystyle\sum_{i=1}^{V} p_i \,\log p_i$ \Comment{Shannon entropy}
\State $H_{\max} \gets \log_2 V$ \Comment{uniform distribution $p_i=\frac1V$}
\State $p_{\text{odd}}\!\gets\!\bigl(H_t/H_{\max}\bigr)^{H_{\text{scale}}}$
\State $\textit{seed}\!\gets\!\mathrm{PRF}(x_{t-1})$
        \Comment{a pseudorandom seed derived from the previous token}
\State $u\!\gets\!\mathrm{Hash2Uniform}(\textit{seed}\oplus K)\in(0,1)$
       \Comment{uniform random variable $u$ derived from a secret key $K$}
\State $g\!\gets\!\mathbf{1}[\,u<p_{\text{odd}}\,]$ \Comment{$g{=}1$: tokens with odd ranks are green.}
\State $\pi\!\gets\!\operatorname*{argsort}(\boldsymbol{\ell}_{t};\downarrow)$
\For{$r=0$ \textbf{to} $V-1$}
  \If{$r\bmod2=g$}
     \State $\boldsymbol{\ell}_{t,\pi[r]}\!\gets\!\boldsymbol{\ell}_{t,\pi[r]}+\delta$
  \EndIf
\EndFor
\State $\hat{x}_{t}\!\gets\!\displaystyle\arg\max_{j}\operatorname{softmax}(\boldsymbol{\ell}_{t})_j$
\State \textbf{return} $\hat{x}_{t}$
\end{algorithmic}
\end{algorithm}

\paragraph{Entropy-driven green-list selection.}
Given \( p_i \),  
the Shannon entropy at time step~\( t \) is defined as:
\begin{equation}\label{eq:shannon_entropy}
  H_t \;=\; -\sum_{i=1}^{V} p_i \,\log p_i,
  \quad
  H_{\max} \;=\; \log_2 V.
\end{equation} 
The entropy is transformed into a Bernoulli parameter  
\begin{equation}\label{eq:podd_def}
  p_{\text{odd}}
  \;=\;
  \left(\frac{H_t}{H_{\max}}\right)^{H_{\text{scale}}},
\end{equation}
where the exponent \(H_{\text{scale}}>0\) controls the steepness of the
mapping.  
Choosing \(H_{\text{scale}}>1\) makes the rise in \(p_{\text{odd}}\)
steeper—almost zero for low-entropy~(sharp) distributions and close to
one only when the distribution becomes flat—thereby protecting fluency
in deterministic contexts.  
Conversely, \(0<H_{\text{scale}}<1\) yields a gentler slope, raising
\(p_{\text{odd}}\) even at moderate entropy and embedding the watermark
more densely when probability mass is already spread across many
tokens.
A uniformly distributed value \(u\), 
deterministically derived from the key,  
is used to select the green parity via $g = \mathbf{1}[\,u < p_{\text{odd}}\,]$.
Under high-entropy conditions, the output distribution of the model 
tends to be relatively uniform, 
indicating that probability mass is distributed across multiple candidate tokens 
and that the second-ranked token is nearly as likely as the top-ranked one. 
In such cases, assigning the odd-ranked group to the green list enhances 
watermarking capacity while maintaining textual fluency. 
In contrast, low-entropy distributions are sharply concentrated on a few top tokens, 
making the even-ranked group a more stable and semantically reliable choice. 
Notably, the odd-ranked group still includes several high-probability candidates 
beyond the top-1 token, allowing watermark insertion without substantially 
degrading generation quality. 
The trade-off between watermark strength and fluency can be further adjusted 
through the entropy scaling parameter~$H_{\text{scale}}$.

\begin{figure*}[hbt!]
    \centering
        \includegraphics[width=0.9\textwidth]{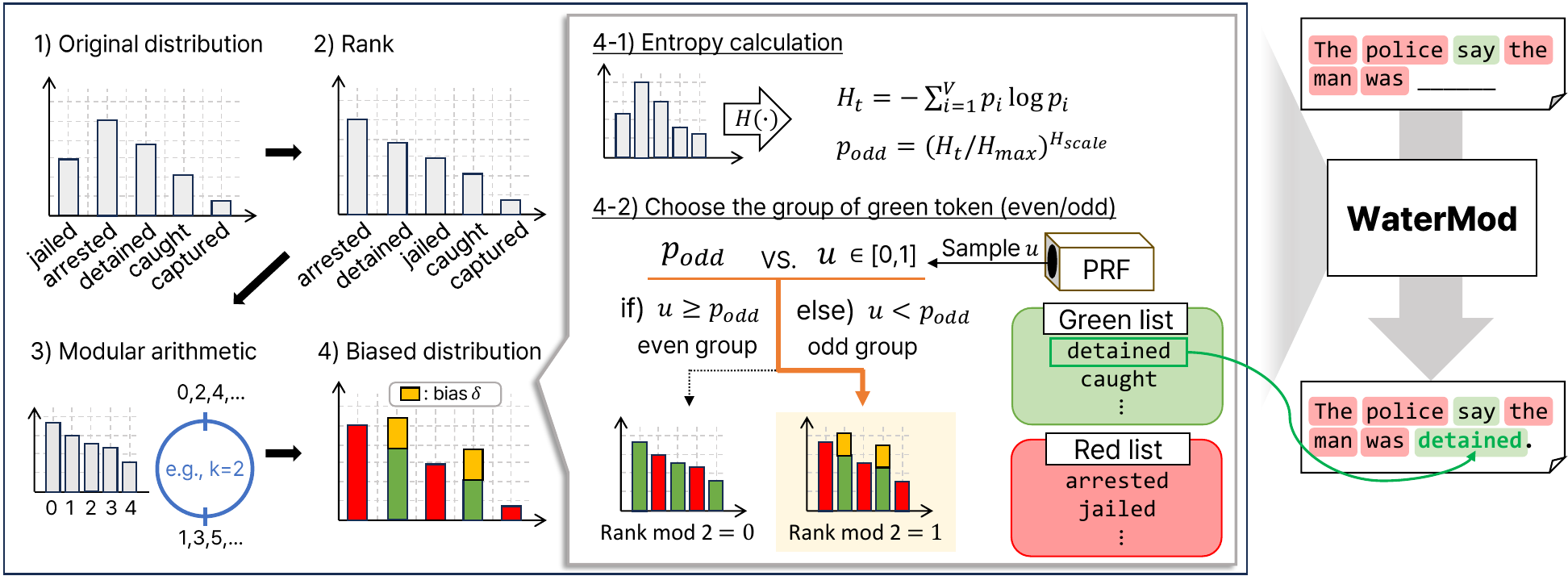}
    \caption{Watermark embedding procedure of WaterMod in the zero-bit setting.
    }\label{fig:overview_zero_bit}
\end{figure*} 

\paragraph{Logit biasing for watermark insertion.}
WaterMod raises every logit whose rank satisfies
\(r\bmod2=g\) by the constant \(\delta\) and 
samples the next token.
The green list is guaranteed to include at least one high-ranked token, 
thereby preserving fluency during generation.
Figure~\ref{fig:overview_zero_bit} illustrates 
the watermark embedding process of WaterMod.

\begin{algorithm}[hbt!]
\caption{\textsc{Zero-bit}\,: Sequence-level Detection}
\label{alg:zerobit_detection}
\small
\textbf{Input:} token sequence $(x_0,\dots,x_{T-1})$, generator $f_\theta$, threshold $\tau$, secret key $K$ \\
\textbf{Output:} \textsc{True} if watermarked, else \textsc{False}
\begin{algorithmic}[1]
\State $G\!\gets\!0$ \Comment{green-token counter}
\For{$t=1$ \textbf{to} $T-1$}
  \State $\boldsymbol{\ell}_{t}\!\gets\!f_\theta(x_{<t})$
  \State recompute $p_{\text{odd}},u,g$ with $\boldsymbol{\ell}_{t},x_{t-1},K$
  \Comment{reconstruction of the green list}
  \State $\pi\!\gets\!\operatorname*{argsort}(\boldsymbol{\ell}_{t};\downarrow)$
  \State $\mathcal{G}\!\gets\!\{\pi[r]\mid r\bmod2=g\}$
  \If{$x_t\!\in\!\mathcal{G}$} \State $G\!\gets\!G+1$ \EndIf
\EndFor
\State $N\!\gets\!T-1$
\State $z\!\gets\!\dfrac{G-\tfrac{N}{2}}{\sqrt{N/4}}$ \Comment{null proportion $\varepsilon=0.5$}
\State \textbf{return} $\mathbf{1}[\,z>\tau\,]$
\end{algorithmic}
\end{algorithm}

\paragraph{$z$-score calculation for watermark detection.}
Algorithm~\ref{alg:zerobit_detection}
outlines the watermark detection procedure under the 
zero-bit watermarking scenario.
The detector reconstructs the green-list parity 
for every position using
the identical secret key,
counts green hits \(G\) over
\(N=T-1\) positions, and evaluates $z$-score:
\begin{equation}\label{eq:zscore_def}
  z=\frac{G-\tfrac{N}{2}}{\sqrt{N/4}}\;.
\end{equation}
A sequence is classified as watermarked when \(z>\tau\).

\subsection{Extension to Multi-bit Watermarking}
\label{sec:multi_bit}

\paragraph{$k$-residue color partition of ranks.}
Algorithm~\ref{alg:multibit_embedding} 
presents the message embedding procedure of WaterMod 
under the multi-bit watermarking scenario.
Applying the modular rule \(r \bmod k\) generalizes the parity split to
\(k\) color classes,
\begin{equation}
  \mathcal{C}_d
  \;=\;
  \bigl\{\,\pi[r] \mid r \bmod k = d \bigr\},
  \quad d \in \{0,\dots,k-1\},
\end{equation}
so every token rank belongs to exactly one residue class \(\mathcal{C}_d\).

\begin{algorithm}[hbt!]
\caption{\textsc{Multi-bit}\,: Embedding a $b$-bit payload at step $t$}
\label{alg:multibit_embedding}
\small
\textbf{Input:} logits $\boldsymbol{\ell}_{t}$, previous token $x_{t-1}$, secret key $K$, bit length $b$, base $k$, payload digits $\mathbf{m}$, bias $\delta$ \\
\textbf{Output:} next token $\hat{x}_{t}$
\begin{algorithmic}[1]
\State $\tilde{b}\!\gets\!\left\lceil b/\log_2 k\right\rceil$              \Comment{length of $\mathbf{m}$ in base-$k$}
\State $\textit{seed}\!\gets\!\mathrm{PRF}(x_{t-1})$
\State $u\!\gets\!\mathrm{Hash2Uniform}(\textit{seed}\oplus K)\in(0,1)$
\State $p\gets\min\!\bigl(\lfloor u\tilde{b}\rfloor,\tilde{b}-1\bigr)$      \Comment{pseudorandom position in the digit string}
\State $d\gets\mathbf{m}[p]$                                                \Comment{digit to embed at this step}
\State $\pi\gets\operatorname*{argsort}(\boldsymbol{\ell}_{t};\downarrow)$  \Comment{rank permutation of the vocabulary}
\For{$r=0$ \textbf{to} $V-1$}                                               \Comment{bias \emph{only} the target color}
   \If{$r\bmod k = d$}
        \State $\boldsymbol{\ell}_{t,\pi[r]}\gets\boldsymbol{\ell}_{t,\pi[r]}+\delta$
   \EndIf
\EndFor
\State $\hat{x}_{t}\gets\arg\max_{j}\operatorname{softmax}(\boldsymbol{\ell}_{t})_j$ 
\State \textbf{return} $\hat{x}_{t}$
\end{algorithmic}
\end{algorithm}

\paragraph{Payload-conditioned color choice.}
A \(b\)-bit message is represented as the base-\(k\) vector:
\begin{equation}
  \mathbf{m}\in\{0,\dots,k-1\}^{\tilde b},
  \quad
  \tilde b=\Bigl\lceil \tfrac{b}{\log_2 k}\Bigr\rceil .
\end{equation}
At step \(t\) the key hash \(u\) selects the pseudorandom position
$p=\min\!\bigl(\lfloor u\tilde b\rfloor,\tilde b-1\bigr),$
and the encoder picks the color
\(d=\mathbf{m}[p]\).

\paragraph{Digit-wise biasing for message embedding.}
WaterMod adds a bias $\delta$ to the logits of all tokens 
satisfying \(r\bmod k=d\). 
Since the probability mass is uniformly distributed across 
color groups, the overall generation quality is preserved.

\begin{algorithm}[hbt!]
\caption{\textsc{Multi-bit}\,: Payload Recovery}
\label{alg:multibit_recovery}
\small
\textbf{Input:} sequence $(x_0,\dots,x_{T-1})$, secret key $K$, generator $f_\theta$, base $k$ \\
\textbf{Output:} recovered digits $\hat{\mathbf{m}}$, $z$-score
\begin{algorithmic}[1]
\State $\tilde{b}\gets|\mathbf{m}|$;\; initialize tallies $C[p][d]\gets0$ \Comment{$C[p][d]$ counts how often color $d$ appears at position $p$}
\State $G\gets0$;\; $T\gets0$                                           \Comment{$G$: hits,\; $T$: inspected steps}
\For{$t=1$ \textbf{to} $T-1$}                              
   \State $\boldsymbol{\ell_t}\gets f_\theta(x_{<t})$                      \Comment{recompute logits for step $t$}
   \State $\textit{seed}\!\gets\!\mathrm{PRF}(x_{t-1})$
   \State $u\!\gets\!\mathrm{Hash2Uniform}(\textit{seed}\oplus K)\in(0,1)$
   \State $p\gets\min\bigl(\lfloor u\tilde{b}\rfloor,\tilde{b}-1\bigr)$  \Comment{digit position used at step $t$}
   \State $\pi\gets\operatorname*{argsort}(\boldsymbol{\ell};\downarrow)$
   \State $r\gets\text{index of }x_t\text{ in }\pi$;\; $d\gets r\bmod k$ \Comment{observed color}
   \State $C[p][d]\gets C[p][d]+1$                                       \Comment{update tallies}
   \If{$d=\mathbf{m}[p]$} \State $G\gets G+1$ \EndIf                     
   \State $T\gets T+1$
\EndFor
\For{$p=0$ \textbf{to} $\tilde{b}-1$}
   \State $\hat{m}[p]\gets\arg\max_d C[p][d]$                            \Comment{majority vote per position}
\EndFor
\State $p_0\gets 1/k$\quad\Comment{null success probability}
\State $z\gets\displaystyle\frac{G-Tp_0}{\sqrt{T\,p_0(1-p_0)}}$          \Comment{$z$-score}
\State \textbf{return} $\hat{\mathbf{m}}$, $z$
\end{algorithmic}
\end{algorithm}

\begin{figure*}[hbt!]
    \centering
        \includegraphics[width=0.90\textwidth]{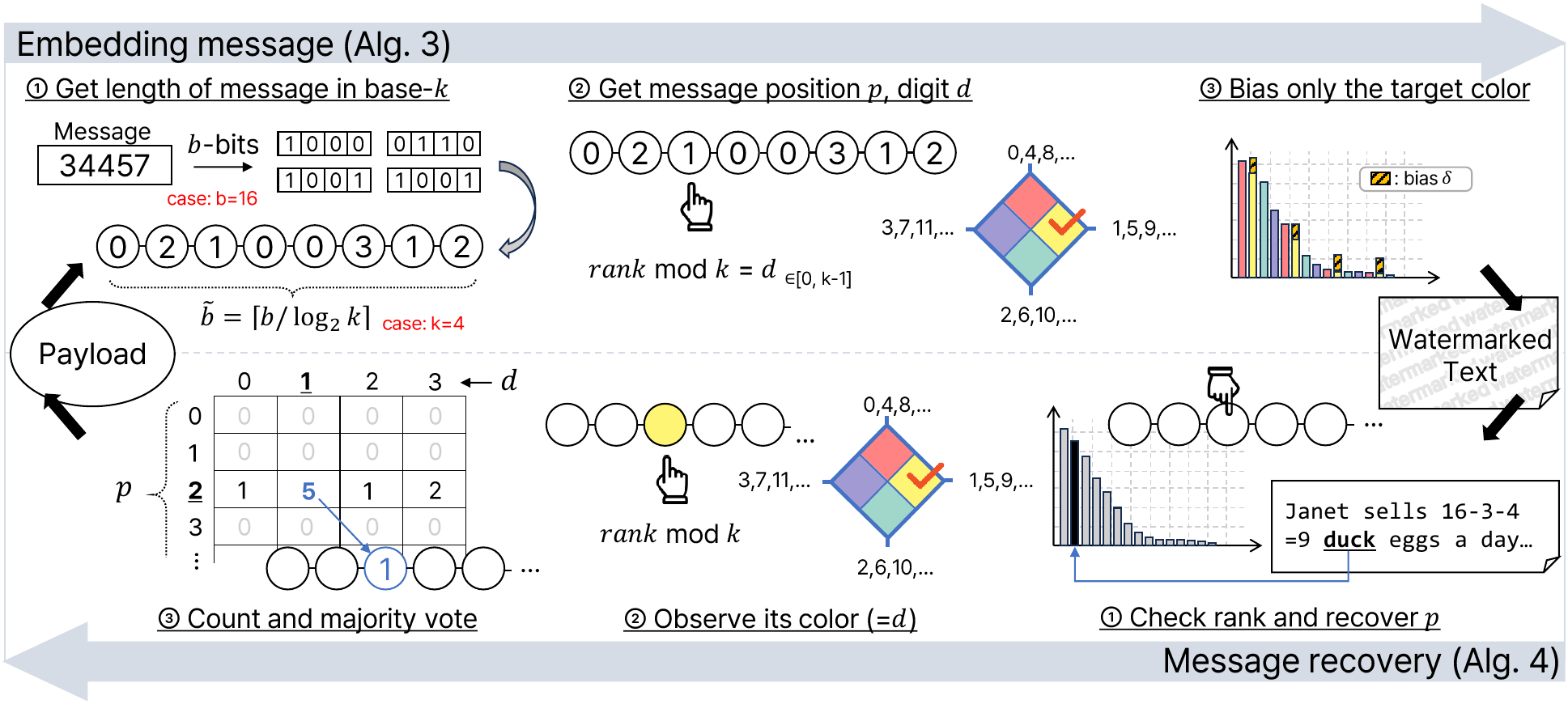}
    \caption{Overview of the message encoding and recovery process in WaterMod 
    under the multi-bit watermarking regime.
    }\label{fig:overview_multi_bit}
\end{figure*} 

\paragraph{$z$-score calculation for watermark detection.}
Algorithm~\ref{alg:multibit_recovery} describes 
the message recovery procedure of WaterMod.
For each token generated after a fixed-length prefix, 
the detector performs the following steps:
1) it reconstructs the target color that should have been favored by the WaterMod encoding scheme;
2) it registers a hit if the observed token belongs to the reconstructed color; and
3) it accumulates the total number of hits \( G \) over \( T \) inspected positions.
Under the null hypothesis---\ie, when no watermark is embedded---each color 
is equally likely to be selected with probability \( p_0 = 1/k \). 
Consequently, the hit count \( G \) follows a binomial distribution:
\begin{equation}\label{eq:g_hit_count}
    G \sim \mathrm{Binom}(T, p_0), \quad
    z = \frac{G - T p_0}{\sqrt{T\,p_0(1 - p_0)}}
\end{equation}
The resulting standardized statistic \( z \) approximately follows 
the standard normal distribution \( \mathcal{N}(0, 1) \) under the null hypothesis. 
In the presence of a watermark, however, the token distribution becomes biased, 
increasing the effective success probability beyond \( p_0 \) and 
shifting \( z \) toward positive values.
A one-sided hypothesis test flags a sequence as watermarked 
if the computed \( z \)-score exceeds a predefined threshold \( \tau \). 
Varying \( \tau \) yields a receiver operating characteristic~(ROC) curve; 
the corresponding area under the curve~(AUROC) quantifies the detection power.
Notably, the same color-position tally table \( C[p][d] \) used for detection 
also supports payload recovery via majority vote. 
Therefore, WaterMod enables both source attribution via the \( z \)-score and 
message retrieval of the embedded digits \( \hat{\mathbf{m}} \) in a single decoding pass.
Figure~\ref{fig:overview_multi_bit} 
provides an overview of the message embedding and recovery processes 
in WaterMod.

\paragraph{Discussion.}
A single modular arithmetic on the probability ranking guarantees
quality preservation in both watermarking regimes.
For \(k=2\) the even–odd split sends near-synonymous tokens to
different sides of the green–red boundary, 
ensuring that every decoding
step retains at least one high-probability candidate.
For \(k>2\) the same mapping distributes the vocabulary almost
uniformly across the \(k\) color lists, 
so multi-bit embedding enjoys
the same fluency safeguard.
Because all algorithms depend on \(k\) only through the residue
condition \(r\bmod k\), 
adjusting that single hyper-parameter moves
WaterMod seamlessly from binary attribution to 
a payload capacity of
\(\log_{2}k\) bits per position.
This unified framework for 
zero- and multi-bit watermarking represents 
the key advance over prior work.


\section{Results and Analysis}
\label{sec:results_and_analysis}

\subsection{Experimental Setup}\label{subsec:results_and_analysis}

\paragraph{Datasets.} We evaluate WaterMod
across three domains: 
\begin{itemize}
    \item \textbf{Natural Language Continuation} We use the news-like subset of the Colossal Common Crawl Cleaned corpus~(C4)~\citep{raffel2020exploring}.
    Given the opening fragment of an article, the model completes the remainder, 
    simulating fake news generation. 
    We randomly sample 500 instances for our experiments.
    \item \textbf{Mathematical Reasoning} We adopt the GSM8K~\citep{cobbe2021training} dataset, 
    which consists of 8,000 arithmetic and grade-school-level math problems 
    designed to assess the reasoning capabilities of LLMs. 
    The task requires solving each problem through 
    chain-of-thought reasoning and presenting the final answer.
    We use the 1,319 instances in the test split.
    \item \textbf{Code Generation} We use the MBPP+~\citep{liu2023evalplus} dataset, 
    where the goal is to generate Python code that 
    satisfies a given problem description 
    written in natural language. 
    MBPP+ comprises 378 programming problems, 
    each accompanied by around 100 test cases, 
    allowing for rigorous functional evaluation of generated code.
\end{itemize}
These three domains exhibit differing levels of 
entropy, 
defined as the entropy of the token probability distribution 
during generation. 
Mathematical reasoning tasks typically exhibit lower entropy 
than natural language generation, 
reflecting the deterministic nature of symbolic computation. 
Likewise, code generation tends to 
produce lower-entropy distributions due to the rigid syntactic 
and structural constraints of programming languages. 
We leverage these variations in entropy across tasks to 
comprehensively assess the performance of WaterMod.
The analysis of entropy differences across tasks is provided in 
Appendix~\ref{sec:appendix_data_entropy}.

\paragraph{Baselines.}
We benchmark WaterMod against five zero-bit schemes 
and one representative multi-bit scheme.
\begin{itemize}
    \item \textbf{KGW}~\citep{kirchenbauer2023watermark} randomly partitions 
    the vocabulary into green/red lists once per step and adds a soft logit bonus 
    to green tokens.
    \item \textbf{EXPEdit \& ITSEdit}~\citep{kuditipudi2024robust} map a pseudorandom seed to a fixed token sequence by exponential minimum or inverse transform sampling.
    \item \textbf{LSH}~\citep{guo2024context} applies 
    locality-sensitive hashing to word embeddings, 
    ensuring that semantically similar tokens are placed in the same green list.
    \item \textbf{SynthID-Text}~\citep{dathathri2024scalable} leverages
    tournament sampling, 
    a sampling technique that subtly aligns token choices with seeded random values.
    \item \textbf{MPAC}~\citep{yoo-etal-2024-advancing} allocates each payload digit 
    to a dedicated token position; 
    at that position the decoder forces a token whose hash 
    equals the digit.
\end{itemize}
EXPEdit, ITSEdit and SynthID-Text represent sampling-based watermarking approaches, 
whereas KGW, LSH, MPAC and WaterMod fall under logit-based watermarking methods.
We evaluate our method and all baselines on the 
same Qwen-2.5-1.5B~\citep{yang2024qwen2} model 
to ensure a fair 
and controlled comparison. 
Since these watermarking approaches, 
including our own, are model-agnostic, 
their principles are generalizable 
to other open-source LLMs.

\paragraph{Evaluation Metrics.}
We evaluate the quality of watermarked outputs using 
task-specific metrics across various datasets. 
On C4, we use perplexity, where lower values indicate more fluent text. 
For GSM8K, we assess accuracy by comparing predicted answers 
to reference solutions. 
On MBPP+, we measure pass@1, 
which calculates the proportion of problems where the 
generated code passes all test cases. 
Detection performance is measured by computing the AUROC from $z$-scores.  
The $z$-scores quantify how likely a text is to have been generated by an LLM.
Higher $z$-scores indicate a stronger likelihood, 
and AUROC evaluates how well the watermarking method distinguishes 
LLM-generated content from human-written text.
Additional details about the evaluation metrics 
are provided in Appendix~\ref{appendix_experimental_settings}.

\paragraph{Implementation Details.}
All methods, including our own and the baseline approaches, use the same configuration.
We apply a deterministic decoding strategy, selecting at each time step the token with the highest probability. 
This eliminates stochastic variability and enables a direct analysis of how watermarking affects 
model outputs.
We limit the maximum number of tokens generated per instance for each dataset 
as follows: 400 for C4, 600 for GSM8K and MBPP+. 
We provide the prompt templates used for each dataset 
in Appendix~\ref{sec:appendix_prompt_templates}.

In the zero-bit watermarking scenario, 
we set the entropy scaling factor \(H_{\text{scale}} = 1.2\) 
when calculating the odd-ranked token selection probability \(p_{\text{odd}}\).
The entropy scaling factor could be further 
optimized according to the entropy level of the domain 
in which the watermark is embedded, 
potentially leading to performance improvements.
We leave the automated discovery of 
the optimal entropy scaling factor for future work.
We configure the watermarking bias $\delta$ as 1.0 for the zero-bit setting 
and 2.5 for the multi-bit setting. 
The higher bias in the multi-bit setting 
is necessary to compensate for its smaller 
target class~($1/k$ of the vocabulary versus $1/2$), 
ensuring sufficient statistical pressure for 
reliable embedding.
For logit-based watermarking methods under the zero-bit setting, 
we fix the green list ratio to 0.5. 
In the multi-bit setting, we embed 16-bit payloads using base $k = 4$.
A 16-bit payload was chosen as it offers 
a practical and expressive message size, 
capable of encoding over 65,000 unique identifiers 
for fine-grained provenance tracing.
We conduct all experiments 
on a single NVIDIA RTX 3090 GPU with 24GB of memory.

\begin{table*}[hbt!]
\centering
\begin{tabular}{l|cc|cc|cc}
\toprule
\multirow{3}{*}{Method} & \multicolumn{2}{c|}{\textbf{C4}} & \multicolumn{2}{c|}{\textbf{GSM8K}} & \multicolumn{2}{c}{\textbf{MBPP+}} \\
\cmidrule(lr){2-3} \cmidrule(lr){4-5} \cmidrule(lr){6-7}
 & Perplexity & AUROC & Accuracy & AUROC & Pass@1 & AUROC \\
\midrule
EXPEdit & 36.35 & 36.90 & 10.84 & 37.37 & 22.80 & 34.38 \\ 
ITSEdit & 31.03 & 11.29 & 11.75 & 35.44 & 20.10 & 27.40 \\ 
KGW & 21.96 & 80.83 & 51.78 & 44.38 & 29.90 & \underline{72.43} \\ 
LSH & 26.19 & \underline{88.03} & \underline{53.07} & 52.63 & \textbf{41.30} & 30.72 \\ 
SynthID-Text & \underline{12.77} & \textbf{94.36} & 47.61 & \underline{97.65} & 27.80 & 66.90 \\ 
WaterMod & \textbf{12.58} & 87.09 & \textbf{53.83} & \textbf{100} & \underline{36.80} & \textbf{82.66} \\ 
\bottomrule
\end{tabular}
\caption{A comparative evaluation of different watermarking methods 
under the zero-bit watermarking scenario. 
We highlight the best-performing result for each evaluation metric in bold, 
and the second-best result with underlining.
}
\label{tab:zerobit_results}
\end{table*}

\subsection{Experimental Results}

\paragraph{Zero-bit Watermarking.}
Table~\ref{tab:zerobit_results} compares the performance of WaterMod 
with five existing watermarking methods 
in the zero-bit watermarking scenario.
On the natural language continuation task~(C4), 
WaterMod achieves the lowest perplexity, 
indicating the most fluent generation among all watermarking methods. 
It ranks third in detection performance~(AUROC 87.09), 
trailing only SynthID-Text and LSH. 
However, the higher AUROC of LSH~(88.03) 
comes at a steep cost in fluency: 
its perplexity is more than twice as high as that of WaterMod.
SynthID-Text emerges as a strong baseline, 
attaining the second-best perplexity and the highest AUROC on C4. 
On the GSM8K mathematical reasoning benchmark, 
WaterMod attains the best accuracy and simultaneously achieves 
perfect detection performance. 
Compared to SynthID-Text, the strongest competing method 
in terms of AUROC, WaterMod improves accuracy by 13.06\%, 
suggesting that the modular operation over token probability ranks 
ensures that the most probable or second most probable token 
is consistently selected, even under watermark embedding.
For code generation on MBPP+, 
WaterMod delivers the best watermark detectability and the 
second-best pass@1 score. 
It outperforms the next-best AUROC baseline, KGW, 
by a substantial 14.12\% while simultaneously improving 
pass@1 by 23.07\%. 
Compared to LSH, which yields the highest pass@1 score 
while exhibiting an extremely weak watermarking signal~(AUROC 30.72),
WaterMod improves AUROC by an impressive 169.07\%, 
indicating that the LSH-based method fails to reliably 
embed the watermark in this low-entropy case.

\begin{table*}[hbt!]
\centering
\begin{tabular}{lcc}
\toprule
\textbf{Source}               & \textbf{Mean $z$-score} & \textbf{AUROC (\%)} \\
\midrule
Human-written text                    & 0.09  & ---   \\
WaterMod~(no attack)          & 14.89 & 100.00 \\
WaterMod~(ChatGPT paraphrase) & 9.95  & 99.95 \\
\bottomrule
\end{tabular}
\caption{Detection robustness on GSM8K 
under a \textsc{ChatGPT} paraphrasing attack.}
\label{tab:paraphrase}
\end{table*}

\begin{table*}[hbt!]
\centering
\begin{tabular}{l|cc|cc|cc}
\toprule
\multirow{3}{*}{Method} & \multicolumn{2}{c|}{\textbf{C4}} & \multicolumn{2}{c|}{\textbf{GSM8K}} & \multicolumn{2}{c}{\textbf{MBPP+}} \\
\cmidrule(lr){2-3} \cmidrule(lr){4-5} \cmidrule(lr){6-7}
 & Perplexity & AUROC & Accuracy & AUROC & Pass@1 & AUROC \\
\midrule

MPAC & 10.88 & 97.78 & 31.77 & 95.05 & 20.60 & 48.40 \\ 
WaterMod & \textbf{10.87} & \textbf{98.02} & \textbf{40.33} & \textbf{96.94} & \textbf{26.20} & \textbf{98.29} \\ 

\bottomrule
\end{tabular}
\caption{A comparative evaluation of different watermarking methods 
under the multi-bit watermarking scenario.
}
\label{tab:multibit_results}
\end{table*}

Practical deployment of watermarking methods requires the embedded signal 
to remain resilient to adversarial rewriting of the text.
We therefore test WaterMod against 
a paraphrasing attack carried
out by ChatGPT~(\texttt{gpt-4o-2025-04-14}).
For every GSM8K sample we supply the WaterMod-marked solution 
to the
assistant with the prompt 
\textit{``Please paraphrase the following text:''}.
The human–written solutions are left untouched, and the same
$z$-score detector is applied to all passages.
Table~\ref{tab:paraphrase} shows that paraphrasing reduces the mean
$z$-score of WaterMod outputs from $14.89$ to $9.95$, 
yet the margin relative to human text~($0.09$) remains large.
The AUROC consequently drops by only
$0.05$ absolute to $99.95$\%.
Because the paraphraser must preserve mathematical correctness, 
many high-rank tokens remain unreplaced.  
As a result, most rank-adjacent alternatives retain their intended color, 
allowing the watermark to survive.  
WaterMod thus maintains near-perfect detection even under strong rewriting attacks.

By default, WaterMod employs Shannon entropy to compute the probability of 
assigning odd-ranked tokens to the green list. 
In Appendix~\ref{sec:appendix_compare_shannon_spike}, 
we evaluate an alternative configuration using spike entropy; 
the results indicate that spike entropy substantially 
improves watermark detection performance, whereas Shannon entropy delivers 
superior task-specific utility.

\paragraph{Multi-bit Watermarking.}
Table~\ref{tab:multibit_results} compares WaterMod against 
MPAC, a recent state-of-the-art multi-bit watermarking approach. 
On all three tasks, 
WaterMod achieves superior performance in both generation quality 
and detection.
In the C4 task, 
WaterMod slightly improves over MPAC in perplexity 
and achieves a higher AUROC, 
underscoring its ability to hide payload bits without degrading fluency. 
In GSM8K, WaterMod significantly enhances accuracy by 
26.94\%, 
while maintaining strong watermark detectability, 
exceeding that of MPAC.
This difference is most pronounced on MBPP+, 
where WaterMod achieves an AUROC of 98.29, 
more than double that of MPAC~(48.40). 
This 103.07\% relative improvement in detectability 
demonstrates that WaterMod remains highly effective even in low-entropy, 
syntactically rigid settings like code. 
Additionally, its pass@1 score improves over 
MPAC by 27.18\%, 
indicating that watermark insertion does not compromise 
program correctness.

\section{Related Work}
\label{sec:related_work} 

LLM-integrated watermarking encodes watermark signals directly within the text generation process.
Early zero-bit approaches introduce subtle biases into the model logits or 
sampling distributions to probabilistically 
flag LLM-generated outputs~\citep{kirchenbauer2023watermark,dathathri2024scalable}.
Subsequent semantic-aware methods improve fluency and detection robustness 
by refining the partitioning of the vocabulary—typically into 
green and red token sets—based on semantic properties~\citep{hou-etal-2024-semstamp}.
More recent multi-bit extensions enable richer payload encoding within model outputs.  
These include bit-string allocation and nested-list biasing strategies, 
as employed in depth watermarking~\citep{yoo-etal-2024-advancing,li-etal-2024-identifying}.
These methods, however, continue to exhibit inherent trade-offs between 
watermark capacity and text quality.
We provide further discussion of related work, 
including post-hoc watermarking, 
in Appendix~\ref{appendix_related_works}.

\section{Conclusion}

WaterMod unifies zero-bit attribution and multi-bit payload embedding 
through a simple $\text{rank}\bmod k$ rule applied to probability-sorted token ranks. 
Experiments across natural language, mathematical reasoning, 
and code generation tasks demonstrate that this probability-balanced bias preserves 
output quality while achieving state-of-the-art detection performance. 

\section*{Acknowledgments}
This work was supported by the NRF grant~(RS-2025-00562134) 
and the AI Graduate School Program~(RS-2020-II201361) 
funded by the Korean government.

\bibliography{aaai2026}

\appendix

\section{Analysis of Average Token Entropy Differences Across Tasks}
\label{sec:appendix_data_entropy}

\begin{figure*}[hbt!]
  \centering
  \begin{subfigure}[b]{0.47\linewidth}
    \centering
    \includegraphics[width=\linewidth]{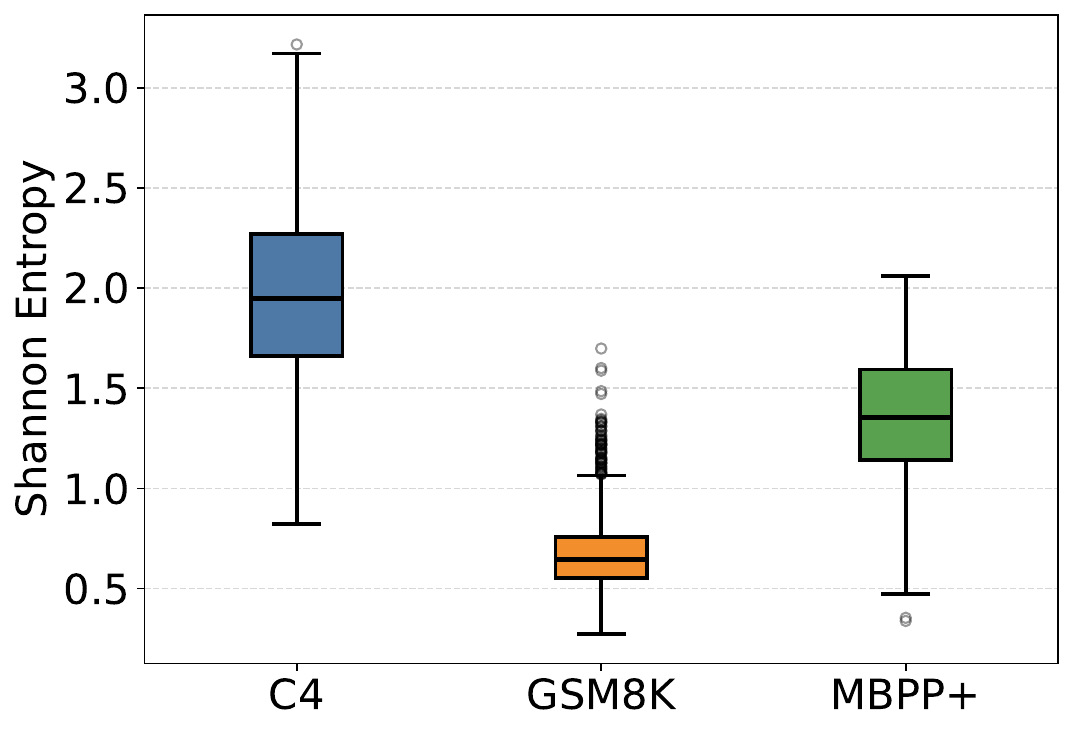}
    \caption{Shannon entropy}
    \label{fig:data_entropy_shannon}
  \end{subfigure}
  \hfill
  \begin{subfigure}[b]{0.47\linewidth}
    \centering
    \includegraphics[width=\linewidth]{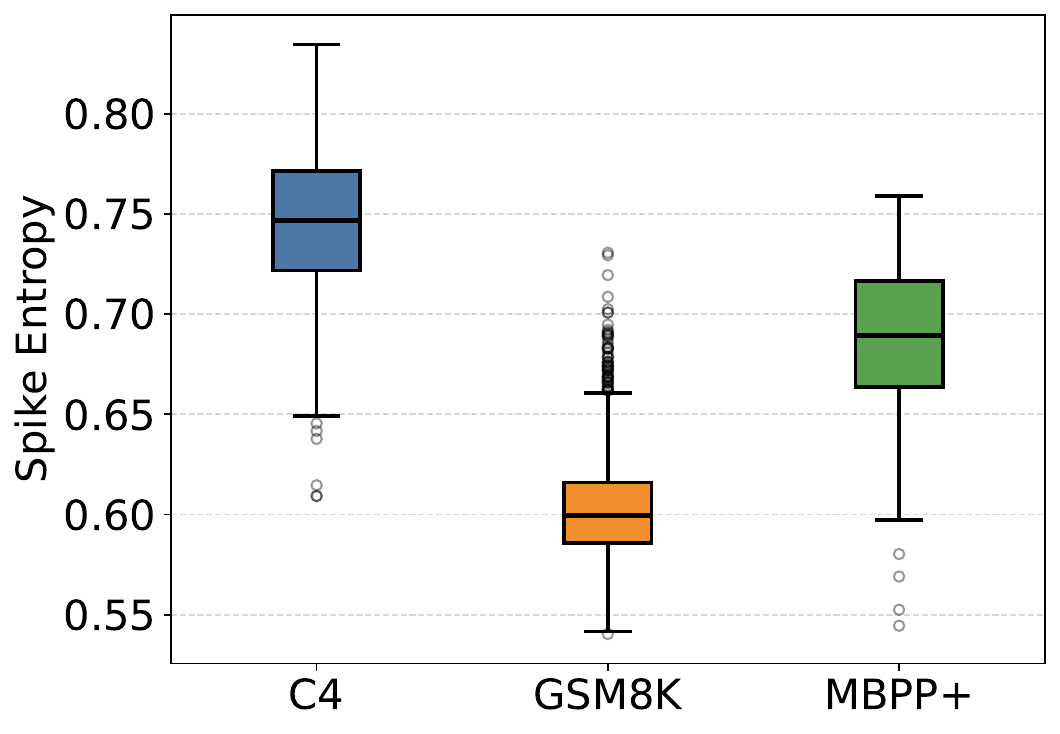}
    \caption{Spike entropy}
    \label{fig:data_entropy_spike}
  \end{subfigure}
  \caption{We report the mean token-level entropy computed over LLM-generated outputs for each dataset. 
  Entropy is measured at every decoding step and averaged across all tokens. 
  Both Shannon entropy~(left) and spike entropy~(right) are presented. 
  Tasks such as mathematical reasoning and code generation show consistently lower entropy than 
  natural language generation, indicating more deterministic token distributions 
  in structured domains.}\label{fig:data_entropy}
\end{figure*}

Figure~\ref{fig:data_entropy} presents the average token-level entropy computed across different datasets, 
offering insight into the variability of output uncertainty across different domains.
We report both the Shannon entropy and spike entropy values. 
For spike entropy, we follow the definition introduced by \citet{kirchenbauer2023watermark}, 
which is defined as follows:
\begin{equation}\label{eq:spike_entropy}
  H_t=\sum_{i=1}^{V}\frac{p_i}{1+\eta p_i},
  \qquad
  H_{\max}=\frac{1}{1+\eta/V}, 
\end{equation}
where $\eta$ is a scalar~(we set it to 1.0). 
We find that mathematical reasoning~(GSM8K) and code generation~(MBPP+) 
involve lower-entropy generation compared to natural language continuation~(C4), 
reflecting the more constrained and deterministic nature of these tasks.
Low-entropy generation environments pose challenges for watermark insertion. 
A low-entropy distribution implies that a single token dominates the probability mass, 
leaving few viable alternatives~(green tokens) for watermarking algorithms to select from. 
As a result, inserting a watermark while preserving content quality becomes significantly more difficult. 
Moreover, effective watermarking relies on introducing detectable 
statistical patterns—such as increased frequency of green tokens. 
In low-entropy settings, 
the limited set of candidate tokens reduces the ability of the model to embed such signals in a reliable manner.

\section{Detailed Experimental Settings}
\label{appendix_experimental_settings}

\paragraph{Evaluation Metrics.}
We adopt task-specific metrics to evaluate the quality of watermarked outputs 
across different datasets. 
On the C4 dataset, we use perplexity. 
A lower perplexity value implies that the language model selects tokens 
with greater confidence, which often correlates with more fluent and coherent text.
On the GSM8K dataset, we evaluate accuracy by comparing predicted answers 
to reference solutions. 
This metric reflects how precisely the model solves mathematical problems. 
For the MBPP+ dataset, we apply pass@1, which measures the proportion of problems for 
which the generated code passes all provided test cases. 

We measure detection performance using the 
AUROC~(Area Under the Receiver Operating Characteristic Curve), based on z-scores. 
We compute a $z$-score for each text by measuring the deviation of 
its watermark score 
from the mean watermark score of human-written text. 
A higher $z$-score indicates a greater likelihood that the text originated 
from an LLM. 
We then calculate AUROC to quantify how effectively the watermarking method 
distinguishes LLM-generated content from human-written text.
We compute the AUROC by comparing the $z$-score distributions of 
human- and LLM-generated texts, 
quantifying how often the LLM text obtains a higher score in pairwise comparisons.

\section{Prompt Templates for LLM-Based Text Generation Across Datasets}
\label{sec:appendix_prompt_templates}

We employ the following prompt templates for LLM-based text generation, tailored to each dataset.

\begin{tcolorbox}[
  enhanced,
  colback=gray!10,
  colframe=black!60,
  boxrule=0.5pt,
  arc=2pt,
  left=4pt, right=4pt, top=4pt, bottom=4pt,
  width=\columnwidth,
  title={\normalsize Prompt Template for C4 Dataset},
  fonttitle=\bfseries,
  coltitle=black,
]
Continue the given text naturally and coherently. Please write only the continuation under \texttt{"Continuation:"}, with no explanation.

\textbf{Given Text:} \texttt{\{context\}}

\textbf{Continuation:}
\end{tcolorbox}

\begin{tcolorbox}[
  enhanced,
  colback=gray!10,
  colframe=black!60,
  boxrule=0.5pt, arc=2pt,
  left=4pt, right=4pt, top=4pt, bottom=4pt,
  width=\columnwidth,
  title={\normalsize Prompt Template for GSM8K Dataset},
  fonttitle=\bfseries,
  coltitle=black,
]
Solve the following math problem step by step. First, provide a detailed step-by-step solution under \texttt{"Solution:"}, then provide only the final answer under \texttt{"Answer:"}, with no explanation.

\textbf{Question:} \texttt{\{question\}}

\textbf{Solution:}

\textbf{Answer:}
\end{tcolorbox}

\begin{tcolorbox}[
  enhanced,
  colback=gray!10,
  colframe=black!60,
  boxrule=0.5pt, arc=2pt,
  left=4pt, right=4pt, top=4pt, bottom=4pt,
  width=\columnwidth,
  title={\normalsize Prompt Template for MBPP+ Dataset},
  fonttitle=\bfseries,
  coltitle=black,
]
Here is a Python programming problem. Implement a function based on the given description. Please write only the code under \texttt{"Code:"}, with no explanation.

\textbf{Problem Description:} \texttt{\{problem\_description\}}

\textbf{Code:}
\end{tcolorbox}

\section{Comparative Evaluation of Shannon and Spike Entropy in WaterMod}
\label{sec:appendix_compare_shannon_spike}

\begin{figure*}[hbt!]
    \centering
        \includegraphics[width=0.85\textwidth]{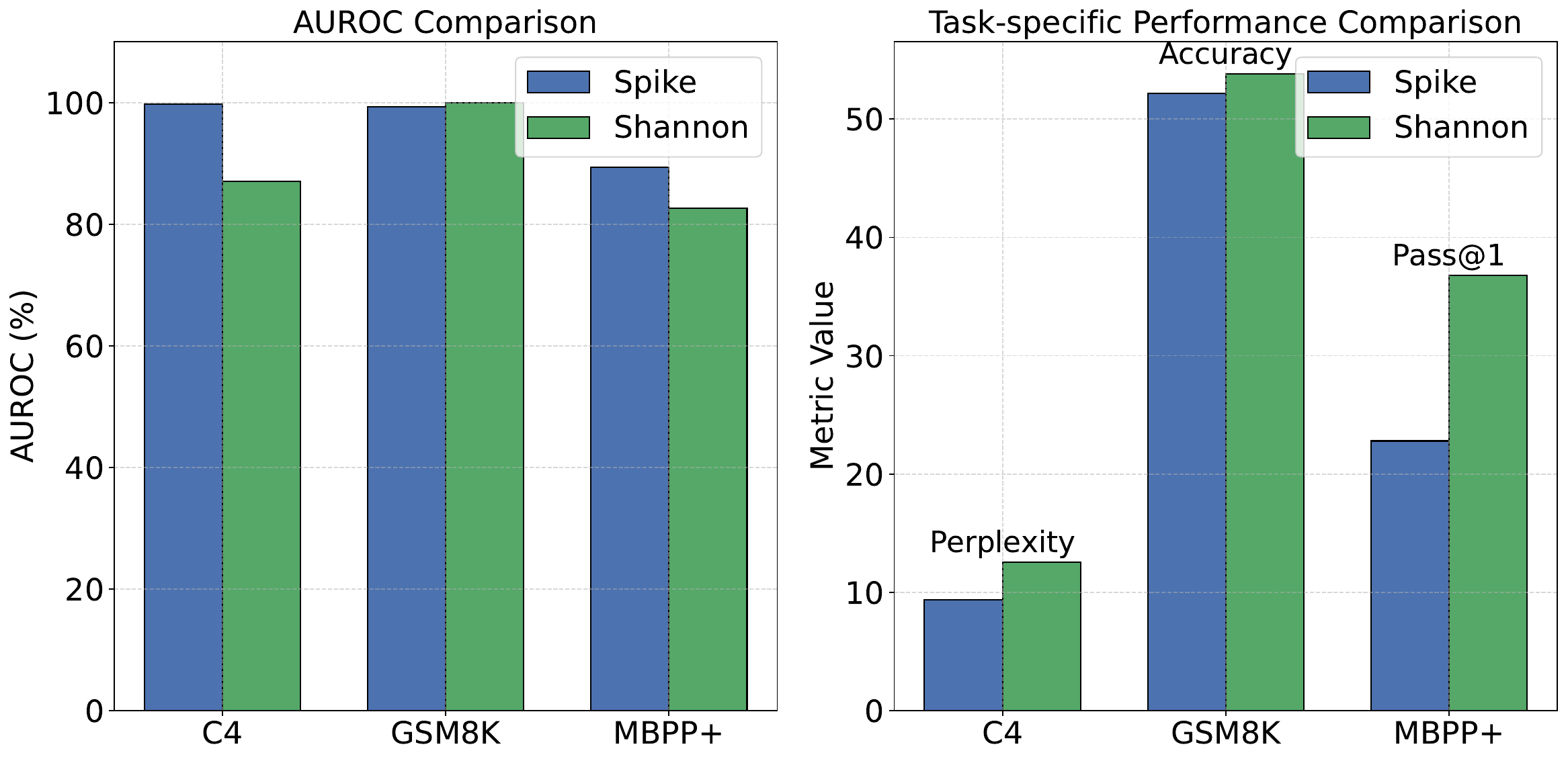}
    \caption{Performance comparison of WaterMod using Shannon entropy and spike entropy 
    under the zero-bit watermarking setting.
    The bar charts on the left present watermark detection performance, 
    while those on the right show task-specific performance. 
    Blue bars represent the performance of WaterMod configured with spike entropy, 
    and green bars correspond to WaterMod using Shannon entropy.
    }\label{fig:entropy_comparison}
\end{figure*}

Figure~\ref{fig:entropy_comparison} compares the performance of 
WaterMod when configured with Shannon entropy versus spike entropy 
under the zero-bit watermarking setting. 
For a detailed explanation of spike entropy, 
refer to Appendix~\ref{sec:appendix_data_entropy}. 
Experimental results show that spike entropy yields better watermark detection performance, 
whereas Shannon entropy offers superior task-specific performance. 
This suggests that spike entropy is preferable in scenarios where detection performance is critical, 
while Shannon entropy is more suitable when maintaining the quality of the watermarked content is prioritized.
Notably, WaterMod demonstrates robust performance regardless of the entropy type, 
indicating that the method can flexibly adapt to different use cases 
by selecting the appropriate entropy formulation.

\section{Further Discussion of Related Work}
\label{appendix_related_works} 

\paragraph{Post-hoc Watermarking.}
Early watermarking approaches operated in a post-hoc fashion by 
modifying text after it had been fully generated.
A completed passage was subtly rewritten to embed an imperceptible 
statistical pattern while preserving its surface form.
Prior studies investigated synonym substitution using lexical resources 
such as WordNet~\citep{Topkara2006hiding} or distributional embeddings like 
Word2Vec~\citep{munyer2023deeptextmark}, 
as well as masked-token infilling with pretrained models~\citep{yang2022tracing}.
Later methods employed public LLM APIs for post-generation synonym 
substitution~\citep{yang2023watermarking}, 
thereby enabling watermarking without altering the model itself.
However, because these methods operated after generation, 
users were able to regenerate unmarked outputs through simple edits.
Similar techniques applied to source code 
corpora~\citep{sun2022coprotector, sun2023codemark} exhibited the same limitations.
These challenges prompted a shift toward model-integrated watermarking, 
in which the signal was embedded directly during decoding.

\paragraph{LLM-Integrated Watermarking.}
Subsequent approaches embedded watermarks directly into the generation process 
by modifying token-level logits or altering the decoding rule.
KGW~\citep{kirchenbauer2023watermark} partitioned the vocabulary 
into green and red token sets and introduced a bias toward the green set during sampling.
\citet{guo2024context} grouped semantically similar tokens to enhance fluency.
Semantic-aware extensions constructed context-sensitive partitions or 
leveraged token redundancy to maintain output quality~\citep{hou-etal-2024-semstamp,hou-etal-2024-k,chen-etal-2024-watme}.

Sampling-based variants avoided logit modification and instead modified the decoding strategy.
SynthID-Text~\citep{dathathri2024scalable} applied tournament sampling in 
production environments, 
while EXP-Edit and ITS-Edit~\citep{kuditipudi2024robust} 
adopted exponential-minimum and inverse-transform sampling, respectively, 
guided by alignment-based detectors.
In the context of code generation, researchers proposed 
grammar-constrained and selective insertion techniques 
to preserve functional correctness~\citep{lee-etal-2024-wrote,guan-etal-2024-codeip}.
Although these zero-bit methods reliably flagged generated content, 
they did not embed explicit payload information.

\paragraph{Multi-bit Extensions.}
Multi-bit watermarking schemes extended these ideas by embedding richer information 
across token positions.
MPAC~\citep{yoo-etal-2024-advancing} introduced logit biases to reflect a designated bit string,
whereas depth watermarking~\citep{li-etal-2024-identifying} targeted nested sublists 
in the vocabulary hierarchy to encode user-specific identifiers.
While these techniques increased capacity, they often suffered from reduced robustness, 
particularly when applied to low-entropy domains such as program code or step-by-step 
mathematical reasoning, where insertion and recovery errors were more likely to occur.

\end{document}